\documentclass[letterpaper, 10 pt, conference]{ieeeconf}  

\IEEEoverridecommandlockouts                              

\usepackage[T1]{fontenc}
\usepackage{cite}
\usepackage{amssymb,amsfonts}
\usepackage{blindtext}
\makeatletter
\let\NAT@parse\undefined
\makeatother
\usepackage{hyperref}
\usepackage{algorithmic}
\usepackage{graphicx}
\usepackage{textcomp}
\usepackage{mathtools}
\usepackage{changes}
\usepackage[font=footnotesize,labelformat=simple]{subcaption}
\usepackage{algorithm}
\usepackage{xcolor}
\usepackage{color, soul}
\usepackage{amsmath}
\usepackage{booktabs}
\usepackage{multirow}
\usepackage{xstring}
\usepackage{hhline}
\usepackage{kotex}
\usepackage{siunitx}
\usepackage{comment}
\usepackage{physics}
\usepackage{tikz}
\usepackage{cleveref}
\usepackage{gensymb}
\usepackage{threeparttable}

\definecolor{pr}{RGB}{0, 0, 0}
\definecolor{gl}{RGB}{0, 0, 0}
\definecolor{v3}{RGB}{0,0,0}
\definecolor{v5}{RGB}{0,0,0}
\definecolor{v7}{RGB}{0,0,255}

\definecolor{rv}{RGB}{0,0,0}
\definecolor{rv1}{RGB}{0,0,0}

\usetikzlibrary{backgrounds}
\pgfdeclarelayer{foreforeground}
\pgfdeclarelayer{foreground}
\pgfdeclarelayer{background}
\pgfsetlayers{background,main,foreground,foreforeground}

\colorlet{mylightred}{red!95!black!30}
\colorlet{mylightblue}{blue!95!black!30}
\colorlet{mylightgreen}{green!95!black!30}
\tikzstyle{policy_nn}=[thick,draw=mylightred!50!black,fill=mylightred,circle,minimum size=11]
\tikzstyle{estimator_nn}=[thick,draw=mylightblue!50!black,fill=mylightblue,circle,minimum size=11]

\title{\LARGE \bf
Energy-Efficient Omnidirectional Locomotion for Wheeled Quadrupeds via Predictive Energy-Aware Nominal Gait Selection
}

\author{Xu Yang$^{1}$, Wei Yang$^{1}$, Kaibo He$^{2}$, Bo Yang$^{1}$, Yanan Sui$^{2}$ and Yilin Mo$^{1\ast}$
\thanks{This work was supported by the National Natural Science Foundation of China under Grants 62461160313, 62273196, 62192752 and the BNRist project (No.BNR2024TD03003).}
\thanks{$^{1}$Department of Automation and BNRist, Tsinghua University, Beijing, China. ({\tt \{yangx21, yang-b21\}@mails.tsinghua.edu.cn, \{rangoyang,ylmo\}@tsinghua.edu.cn})}%
\thanks{$^{2}$School of Aerospace Engineering, Tsinghua University, Beijing, China. ({\tt hkb21@mails.tsinghua.edu.cn, ysui@tsinghua.edu.cn})}\hfill \break%
\thanks{*Corresponding author}\hfill \break%
}

\begin{document}
\maketitle
\thispagestyle{empty}
\pagestyle{empty}

\begin{abstract}
    Wheeled-legged robots combine the efficiency of wheels with the versatility of legs, but face significant energy optimization challenges when navigating diverse environments. In this work, we present a hierarchical control framework that integrates predictive power modeling with residual reinforcement learning to optimize omnidirectional locomotion efficiency for wheeled quadrupedal robots. Our approach employs a novel power prediction network that forecasts energy consumption across different gait patterns over a 1-second horizon, enabling intelligent selection of the most energy-efficient nominal gait. A reinforcement learning policy then generates residual adjustments to this nominal gait, fine-tuning the robot's actions to balance energy efficiency with performance objectives. Comparative analysis shows our method reduces energy consumption by up to 35\% compared to fixed-gait approaches while maintaining comparable velocity tracking performance. We validate our framework through extensive simulations and real-world experiments on a modified Unitree Go1 platform, demonstrating robust performance even under external disturbances. Videos and implementation details are available at \href{https://sites.google.com/view/switching-wpg}{https://sites.google.com/view/switching-wpg}.
\end{abstract}

\section{Introduction}
    Efficient and agile locomotion is a key objective in the motion control domain of mobile robotics. In recent years, wheeled-legged robots~\cite{dietrich2016whole, klamt2020remote, bjelonic2019keep, bjelonic2021whole, bjelonic2022survey} have emerged as a promising solution that combines the speed and energy efficiency of wheels along with the maneuverability of legs, offering the potential for both efficient and versatile locomotion across diverse environments. However, optimizing energy consumption across diverse gaits remains a significant challenge for these hybrid platforms, requiring intelligent management of the robot's dual locomotion systems. The energy consumption profile varies dramatically between different locomotion strategies—pure wheeled motion typically consumes minimal energy on flat surfaces, while various legged gaits offer different trade-offs between stability, maneuverability, and power efficiency. Making optimal decisions about which gait to employ and when to transition between gaits significantly impacts the overall energy efficiency and operational duration of these robots.

    The challenge is further compounded by the mechanical complexity of wheeled-legged platforms. To reduce system complexity and cost, many designs (including our platform) use single unsteerable wheels at each leg~\cite{de2019trajectory, bjelonic2019keep, bjelonic2020rolling}. This design choice significantly increases the likelihood of tire sideslip during turning maneuvers, resulting in complex frictional interactions that are difficult to model accurately. These undesirable slipping behaviors not only complicate controller design but also lead to substantial energy losses through friction, further emphasizing the need for energy-aware locomotion strategies in these systems.

    In this work, we address these challenges by developing a predictive energy-aware framework that enables wheeled-legged robots to proactively select energy-optimal gaits. Our approach integrates a neural network that forecasts power consumption across different gait patterns, allowing the robot to make informed decisions before transitions become necessary. By implementing a curriculum learning strategy that gradually shifts focus toward energy efficiency, our framework learns to balance the competing objectives of energy minimization, locomotion stability, and command tracking accuracy.

    \subsection{Related Work}
    \begin{figure}[t!]
        \centering
        \includegraphics[width=0.48\textwidth]{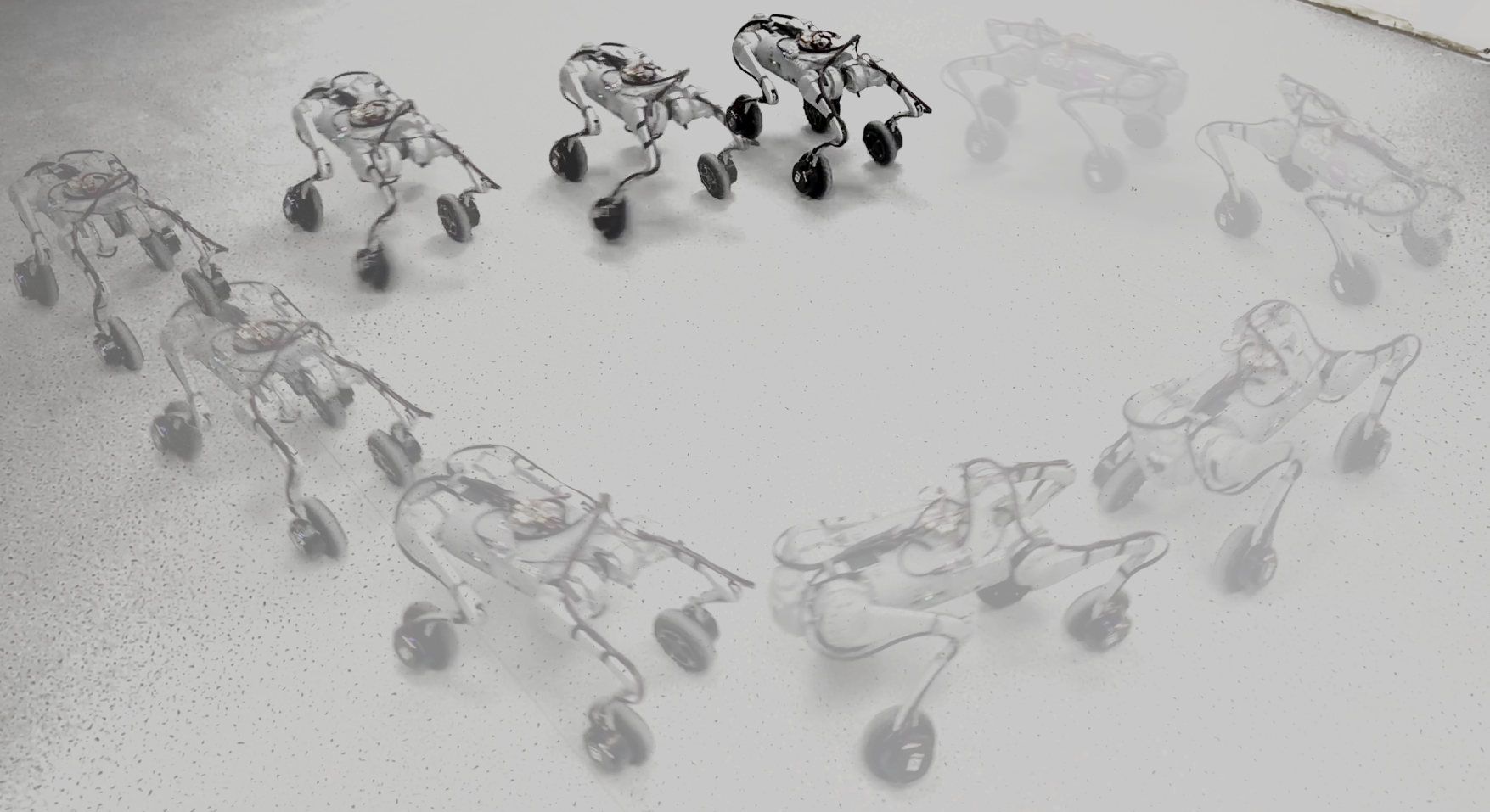}
        \captionsetup{font=footnotesize}
        \caption{Real-world experiment on omnidirectional locomotion: forward, turning and lateral movement of our wheeled quadrupedal robot.}
        \label{fig:cover}
    \end{figure}

    Recent advancements in large-scale parallel simulation engines have significantly accelerated RL-based control development for legged locomotion~\cite{makoviychuk2021isaac, rudin2022learning}. RL policies has demonstrated unprecedented ability to produce rapid and robust legged motion control~\cite{hwangbo2019learning, kumar2021rma, margolisyang2022rapid, fu2022coupling}. However, despite these impressive capabilities, this approach often results in aggressive policies that prioritize task completion over energy efficiency. Balancing these competing objectives remains a significant challenge for end-to-end learning methods.

    To address this inherent limitation, prior works have explored various strategies for quadrupedal locomotion optimization. These methodologies can be broadly categorized into two approaches: predefined gait sequence integration and handcrafted reward shaping. The first approach, exemplified by Central Pattern Generators (CPGs)~\cite{shao2021learning, shi2022reinforcement,lee2022control} and Policies Modulating Trajectory Generators (PMTGs)~\cite{iscen2018policies, lee2020learning}, directly incorporates predefined gait sequences as nominal actions into RL frameworks. This integration enables robots to adapt their gait patterns based on environmental conditions and robot state. However, the nominal gait selection is typically done manually~\cite{lee2020learning,lee2022control, margolis2023walk}, limiting the adaptability and energy efficiency of the resulting policies.

    In contrast, the second approach employs handcrafted reward functions within end-to-end learning frameworks that guide the optimization process towards energy-efficient solutions by explicitly penalizing high-torque or high-power actions~\cite{Fu2021MinimizingEC, liang24adaptive, Shafiee2024ViabilityLT, lee24wheel}. These methods learn policies from scratch without any predefined motion primitives or nominal gaits as guidance. Nevertheless, a fundamental challenge persists: the effective fusion of energy-efficiency rewards with task-specific objectives within a single reward function. This difficulty arises because these dual objectives frequently conflict with each other, creating an inherent tension in the optimization process. Moreover, the training process often requires significantly more time and data to converge due to the absence of prior motion knowledge.

    In this work, we develop a hierarchical control framework tailored for wheeled quadrupedal robots, enabling energy-efficient tracking of commanded velocities. With this framework, the robots can autonomously transition between various gaits while maintaining high stability and low energy consumption.

    \subsection{Contributions}
    The main contributions of the paper are summarized as follows:
    \begin{itemize}
        \item We introduce a hierarchical control framework that integrates predictive energy-aware gait selection with residual reinforcement learning. This approach optimizes omnidirectional locomotion efficiency for wheeled quadrupedal robots by forecasting energy consumption across different gait patterns and selecting the most efficient nominal gait before applying RL-generated refinements.
        \item We develop an augmented gait library that integrates wheeled velocities with traditional legged locomotion patterns in prior works. This integration enables more efficient and versatile omnidirectional movement capabilities specifically tailored for wheeled quadrupedal robots.
        \item We validate the effectiveness of our proposed framework through comprehensive evaluation in both simulation and real-world experiments on a modified Unitree Go1 platform. Our method achieves significant reduction in energy consumption compared to end-to-end and fixed-gait approaches while maintaining comparable velocity tracking performance, even under external disturbances.
    \end{itemize}

    \section{Hardware}\label{sec:hardware}
\begin{figure}[!t]
        \centering
        \vspace{0.2cm}
        \input{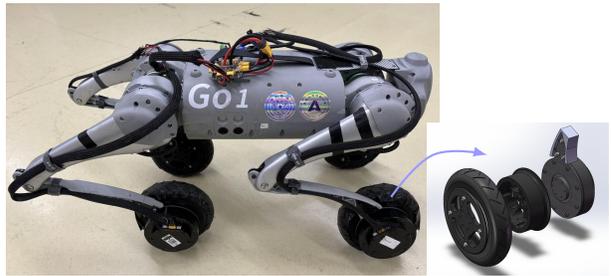}
        \captionsetup{font=footnotesize}
        \caption{Our low-cost wheeled-legged robot and the exploded view of the modified structure. We have open-sourced our hardware design and firmware code, which is available for download from the project site.}
        \label{fig:hardware}
    \end{figure}

    We first introduce our low-cost wheeled-legged quadruped platform, which is modified from an existing commercial robot. As depicted in Fig.~\ref{fig:hardware}, our prototype is built upon the Unitree Go1 quadruped robot~\cite{unitree2024}, with a total weight of 16 kg. For each leg, we replace the standard foot with a rubber tire and incorporate a Tmotor AK80-6 motor to drive the wheel. This modification results in a robot possessing a total of 16 degrees of freedom (DoF): 12 DoF from the original leg joints and 4 additional DoF from the wheel actuators. Additionally, the platform includes an Inertial Measurement Unit (IMU) for orientation and acceleration sensing, and each motor is equipped with an integrated encoder for precise movement tracking.
    \vspace{0.3cm}

    \section{Proposed Method}\label{sec:method}
    \begin{figure*}[tbhp]
        \vspace{0.3cm}
        \centering
        \input{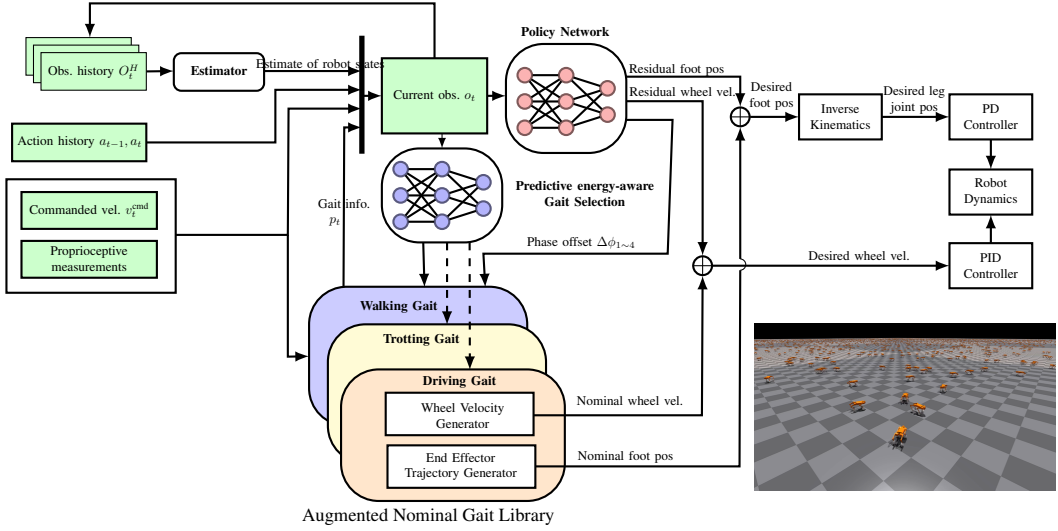}
        \captionsetup{font=footnotesize}
        \caption{Overview of the proposed framework: By predicting energy consumption for each gait, the augmented nominal gait with highest energy efficiency is selected from the gait library, generating both wheel velocities and desired end-effector positions for the legs. A policy network, trained in simulation, compensates for the limitations of the nominal gait. It takes as input the history of actions, estimated robot states $z_t$, user commands $v_t^{\text{cmd}}$, proprioceptive measurements, and augmented nominal gait information $p_t$. The network then outputs residuals for the desired end-effector positions of the legs, wheel velocities, and phase adjustments. Together, the nominal gait and the learned residual policy determine the robot's movements.}
        \label{fig:framework}
    \end{figure*}
    Fig.~\ref{fig:framework} illustrates the proposed hierarchical control architecture, which consists of an energy consumption prediction module for nominal gait selection and a learned residual RL policy. The nominal gait, computed based on the robot's state and commanded velocities, integrates periodical legged movements with wheel velocities to leverage the hybrid mobility of wheeled-legged robots. This primitive motion is then refined by a policy network trained in massive parallel simulations, enabling the robot to adapt to various scenarios and enhancing resilience to unexpected disturbances. This hierarchical approach allows the robot to achieve both high stability and energy efficiency, addressing the unique challenges of controlling robots with both legged and wheeled locomotion capabilities.

    \subsection{Augmented Nominal Gait Design}
    The nominal gait fundamentally determines the locomotion pattern, significantly influencing the robot's overall performance. Building upon previous research on quadrupedal gaits~\cite{lee2020learning, lee2022control}, we have developed a nominal gait library that includes driving, walking, and trotting gaits, designed for easy modification and expansion. To parameterize gait patterns for wheeled-legged quadrupeds, we adopt a phase characterization for each leg's gait cycle, denoted as $\phi_{1,2,3,4}^t\in[0,1)$, which updates dynamically based on gait frequency $f$ and control time step $dt$:
    \begin{equation}
        \begin{aligned}
            \phi_{i}^{t+1}=(\phi_{i}^{t}+f\cdot dt )(\operatorname{mod}\ 1), i=1,2,3,4.
        \end{aligned}
    \end{equation}

    In our framework, each leg's periodical movement is defined by the duty factor $\text{DF}_{\text{swing}}\in[0,1]$, representing the proportion of swing phase within the full gait cycle. During the stance phase ($\phi_{i}^t\in[\text{DF}_{\text{swing}}, 1)$), the wheel's position remains fixed relative to the hip link frame, determined solely by body height $h_{\text{b}}$. During the swing phase, the leg trajectory incorporates stepping height $h_{\text{s}}$ and follows two smooth curves. With normalized swing progress $\psi\triangleq \phi_{t}/\text{DF}_{\text{swing}}$ (ranging from 0 to 1), the swing trajectory relative to the hip frame is defined by lift-off and touch-down cubic Hermite segments:
    \begin{equation}
        F(\psi)\!=\! \begin{cases}
            \!-h_b\!+\!h_{\text{s}}(-16\psi^3\!+\!12\psi^2) & \!\psi \!\in\! [0, 0.5],\\
            \!-h_b\!+\!h_{\text{s}}(16\psi^3\!-\!36\psi^2\!+\!24\psi\!-\!4) & \!\psi \!\in\! [0.5,1].
        \end{cases}
    \end{equation}

\begin{figure}[!t]
        \vspace{0.2cm}
        \centering
        \input{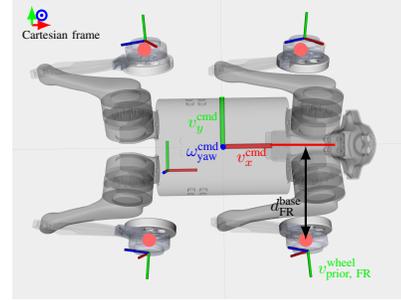}
        \captionsetup{font=footnotesize}
        \caption{Top view of our wheeled quadrupedal robot. The commanded velocity $v_x^{\text{cmd}}, v_y^{\text{cmd}}$ and $\omega^{\text{cmd}}$ is shown in the robot's base frame.}
        \label{fig:sketch}
    \end{figure}

    An innovation in our framework is the integration of wheel velocity into the nominal quadrupedal gait. By considering the wheeled quadrupedal robot as an individual-wheel-driven vehicle, the nominal wheel velocity can be computed based on the commanded velocity $v_t^{\text{cmd}}$ and the robot's leg configuration $q_t^{\text{leg}}$. User commands $v_t^{\text{cmd}}\in\mathbb{R}^3$ include base forward velocity $v_{x, t}^{\text{cmd}}$, lateral velocity $v_{y, t}^{\text{cmd}}$, and yaw rate $\omega_t^{\text{cmd}}$. As shown in Fig.~\ref{fig:sketch}, the nominal velocity of the front-right (FR) wheel when in ground contact is given by:
    \begin{equation}
        v_{\text{n}, \text{FR}}^{\text{wheel}}=-(v_{x}^{\text{cmd}}-d_{\text{FR}}^{\text{base}}\omega^{\text{cmd}}) / R_{\text{wheel}},
    \end{equation}
    where $d_{\text{FR}}^{\text{base}}$ represents the distance from the base to the FR wheel along the robot's $y$ axis and $R_{\text{wheel}}$ is the wheel radius. Without ground contact, wheel velocity is preserved from the previous moment to conserve energy. The nominal velocities for the remaining wheels are calculated in a similar way.

    \begin{table}[!t]
        \centering
        \captionsetup{font=footnotesize}
        \caption{Nominal gait parameters.}
        \label{table:gait_library}
        \begin{center}
        \begin{tabular}{cccc}
        \hline
        \hline
        & Driving & Trotting & Walking\\
        \hline
        $\phi_{1\sim4}$ & [0, 0, 0, 0] & [0, 0.5, 0, 0.5] & [0, 0.25, 0.5, 0.75]\\
        $\text{DF}_{\text{swing}}$ & 0 & 0.4 & 0.225\\
        $f$ & 0 & 1.2 & 0.8\\
        \hline
        \hline
        \end{tabular}
        \end{center}
    \end{table}

    The specific parameters for each gait are detailed in Table~\ref{table:gait_library}. The ability to dynamically select the most energy efficient gait primitive is crucial for both accelerating training and improving the robot's energy efficiency. In this paper, the nominal gait is selected with the prediction of energy consumption in a short horizon, as described in the following subsection.

    \subsection{Predictive Energy-Aware Nominal Gait Selection}
    To predict the energy consumption of different gait primitives, we develop a power prediction network that forecasts the power consumption of each gait pattern over a 1s horizon (roughly a single gait cycle). The network takes as input the robot's observation $o_t$ and outputs the predicted power consumption $p_{t, \text{est}}\in\mathbb{R}^{3}$ for each gait. The predicted power consumption is then used to select the most energy-efficient gait for the robot's current state. 
    
    The nominal gait is selected by randomly sampling from the gait library with the probabilities proportional to $\exp(-p_{t, \text{est, i}}/\tau)/\sum_i\exp(-p_{t, \text{est, i}}/\tau)$, where $i$ indicates the gait type and $\tau$ is a temperature parameter that controls the exploration-exploitation trade-off. The sampling strategy ensuring that the robot's gait selection is biased towards more energy-efficient gaits. 

    The power prediction network is trained in simulation along with the residual RL policy. In order to facilitate the training process, we adopt a curriculum learning strategy that gradually shifts the focus from exploration towards energy efficiency. The curriculum learning strategy is implemented by annealing the temperature parameter $\tau$ over the training iterations, starting from a high value that encourages exploration and gradually decreasing to a lower value that emphasizes exploitation. This strategy enables the robot to learn energy-efficient locomotion strategies while maintaining high exploration rates during the early stages of training.

    \subsection{Residual RL Policy}
    For blind locomotion control of wheeled quadrupedal robots, we define the state of an agent as $s_t\triangleq \langle o_t, x_t\rangle$, where $o_t$ represents the realistically accessible observation vector and $x_t$ is the privileged information. While $x_t$ is typically unavailable in real-world scenarios, it can be accessed in simulations. Additionally, we define an observation history of length $H$ as $O_t^H\triangleq \langle o_t, o_{t-1}, \cdots, o_{t-H+1}\rangle$. In this work, we use a history length $H=6$ for our robot.

    \textbf{Observation Space of the Actor}: In our setting, the observation of the actor $o_t$ consists of proprioceptive sensor measurements, augmented nominal gait information, estimated robot states, user commands and the previous actions.
        \begin{enumerate}
        \item \textbf{Sensor measurements} include angular velocity $\omega_t\in\mathbb{R}^3$, base orientation $g_t\in\mathbb{R}^3$, the robot's leg configuration $q_{t}^{\text{leg}}\in\mathbb{R}^{12}$ and the joint velocity $\dot{q}_t\in\mathbb{R}^{16}$. These measurements are obtained with an IMU and all joint encoders.
        \item The \textbf{augmented nominal gait state} $p_t\in\mathbb{R}^{18}$ is described by each leg's phase, gait frequency, duty factor and nominal wheel velocities, expressed as
        \begin{equation}
            \begin{aligned}
            [\sin(2\pi\phi_{1\sim 4}^t), \cos(2\pi\phi_{1\sim 4}^t), \\
            \Delta \phi_{1\sim 4}, f, \text{DF}_{\text{swing}}, v_\text{n}^{\text{wheel}}].
            \end{aligned}
        \end{equation}
        \item \textbf{Estimated robot states} $z_t$ include the planar linear velocity $v_{xy,t}^{\text{base}}\in\mathbb{R}^{2}$ and the base height $h_{t}^{\text{base}}\in\mathbb{R}$, calculated with the observation history $O_t^H$ using a filter or a neural network trained to be an estimator. Further details are provided in the {\textbf{Estimator Module}} part.
        \item \textbf{User commands} $v_t^{\text{cmd}}\in\mathbb{R}^3=[v_{x, t}^{\text{cmd}}, v_{y, t}^{\text{cmd}}, \omega_{t}^{\text{cmd}}]$.
        \item \textbf{Previous actions} from the past two time steps $a_{t-1}, a_{t-2}\in\mathbb{R}^{22}$, which are included to ensure smooth control inputs.
        \end{enumerate}
        Thus, the actor's full observation is defined as
        \begin{equation}
        o_t = \langle\omega_t, g_t, q_{\text{leg}, t}, \dot{q}_t, p_t, z_t, v_t^{\text{cmd}}, a_{t-1},a_{t-2}\rangle.
        \end{equation}

    \textbf{Action Space of the Actor}: The action $a_t$ consists of phase residuals $\Delta \phi_{1\sim 4}\in\mathbb{R}^4$, end effector position residuals for the legs $\Delta p_{t}^{\text{ee}}\in\mathbb{R}^{12}$ and wheel velocity residuals $\Delta v_{t}^{\text{wheel}}\in\mathbb{R}^4$. To preserve the nominal gait pattern while leveraging the robustness of the learned policy, both learned and nominal actions are integrated to determine the robot's behavior. The desired end effector position for the legs is computed by $p_{\text{des},t}^{\text{ee}} = p_{\text{n},t}^{\text{ee}} + \Delta p_{t}^\text{ee}$, which is then converted into desired joint angles $q_{\text{des}, t}^{\text{leg}}$ via inverse kinematics. These angles are tracked by proportional-derivative (PD) controllers. Similarly, the desired wheel velocity $v_{\text{des}, t}^{\text{wheel}}=v_{\text{n}, t}^{\text{wheel}} + \Delta v_{t}^{\text{wheel}}$ is regulated by a proportional-integral-derivative (PID) controller. Although a PD controller could also be used for wheel velocity control, we find that the PID controller offers better stability and accuracy due to the complex friction dynamics involved.

    \textbf{Observation Space of the Critic}: The critic is used solely during training and is not required in deployment. As a result, its observation space can be augmented to include privileged information $x_t$, such as ground friction, restitution, wheel contact forces, disturbance forces. Incorporating this additional privileged information enhances the critic's performance, thereby accelerating the training process.
    
    {\bf{Reward Function:}} The reward function plays a critical role in guiding the behavior of agents. The specific reward terms are listed in Table~\ref{table_reward}.

    {\bf{Estimator Module:}} The observation history $O_t^H$ can be utilized for estimating the robot's state, identifying physical parameters or inferring environmental information. In this work, we use a multi-layer perceptron (MLP) as a state estimator. This MLP is trained via supervised learning in simulations and subsequently deployed in the real world to estimate the robot's planar linear velocity $v_{xy,t}^{\text{base}}\in\mathbb{R}^2$ and base height $h_{t}^{\text{base}}\in\mathbb{R}$.

     \begin{table}[!t]
        \centering
        \captionsetup{font=footnotesize}
        \caption{Setting of reward functions. The notation $(\cdot)^{\text{cmd}}$ signifies the commanded value, and the terms $x,y$ and $z$ are defined within the frame of the robot's body. The variables $g_{xy}, \omega_{xy}$, $\tau$ and $n_{\text{collision}}$ represent the orientation's $xy$ component, angular velocity's $xy$ component, joint torques and the number of collisions on parts of the robot's body, respectively.}
        \label{table_reward}
        \begin{center}
            \begin{tabular}{ccc}
                \hline
                \hline
                Reward Term & Definition ($r_i$) & Weight ($w_i$) \\
                \hline
                Forward velocity tracking & $\exp\{-4(v_x^{\text{cmd}}-v_x)^2\}$ & 2\\
                Lateral velocity tracking & $\exp\{-4(v_y^{\text{cmd}}-v_y)^2\}$ & 2\\
                Angular velocity tracking & $\exp\{-4(\omega^{\text{cmd}}-\omega)^2\}$ & 3\\
                Residual penalty & $\|a_t\|$ & -0.3\\
                Orientation & $\|g_{xy}\|^2$ & -2\\
                Linear velocity ($z$) & $v_z^2$ & -2\\
                Angular velocity ($xy$) & $\|\omega_{xy}\|^2$ & -0.05\\
                Joint accelerations & $\|\ddot q\|^2$ & $-2.5\times 10^{-7}$ \\
                Joint torques & $\|\tau\|^2$ & $-10^{-5}$ \\
                Action smoothness& $(a_t+a_{t-2}-2a_{t-1})^2$ & -0.01 \\
                Collision & $n_{\text{collision}}$ & -2\\
                Energy Efficiency & $\|\tau\cdot\dot q\|^2$ & $-1\times 10^{-5}$\\
                \hline
                \hline
             \end{tabular}
        \end{center}
        \end{table}

    \section{Evaluation}\label{sec:training}

    We conducted the training of wheeled quadrupedal robots on flat ground using the proposed RL framework within a simulated environment. The robot model is a modification of the Unitree Go1 quadruped robot. The simulation environment is built in the IsaacGym simulator~\cite{makoviychuk2021isaac} based on the legged-gym repository~\cite{rudin2022learning}. The training is executed using a single NVIDIA RTX 4090 GPU, accommodating 4096 robots in parallel. The training process spanned approximately 3 hours in real-time. We set the maximum episode duration to 20 seconds, with episodes terminating either upon the robot's base contacting the ground. The simulation time step $dt$ is set to 0.001 seconds to ensure accurate dynamics simulation, and the RL policy along with the predictive module operates at a frequency of 50Hz.

    To evaluate the effectiveness of our framework, we compare the following methods:
    \begin{enumerate}
        \item {\bf{Baseline}}: In this approach, the robot operates without any predefined gait primitive or predictive module, and the policy is entirely trained from scratch.
        \item {\bf{PMTG}}~\cite{lee2022control}: This method uses a fixed trotting gait for nominal leg pattern, with the robot's wheel velocities determined by RL policy. Unlike our approach where the gait switching is determined by predicted energy consumption, the switching is done fully by the residual RL policy.
        \item {\bf{Ours w/o nominal wheel velocity}}: This variant of our framework employs a switching nominal gait based on predictive power consumption, without nominal wheel velocities.
        \item {\bf{Ours w/o predictive module}}: In this method, we use a fixed augmented trotting gait as a gait primitive, which is distinguished from the PMTG method by the inclusion of wheel velocities.
        \item {\bf{Ours}}: Our full proposed framework with automatic switching of the augmented nominal gait with a predictive module.
    \end{enumerate}

    All methods are trained with the same set of reward functions outlined in Table~\ref{table_reward}, with an exception for the baseline method. Due to the absence of a nominal gait, the baseline method omits the residual penalty from its reward function. For fair comparison, we set all training hyper parameters to be consistent across all methods and the training lasts for 4000 epochs.
    \begin{figure}[!t]
        \centering
        \includegraphics[width=0.45\textwidth]{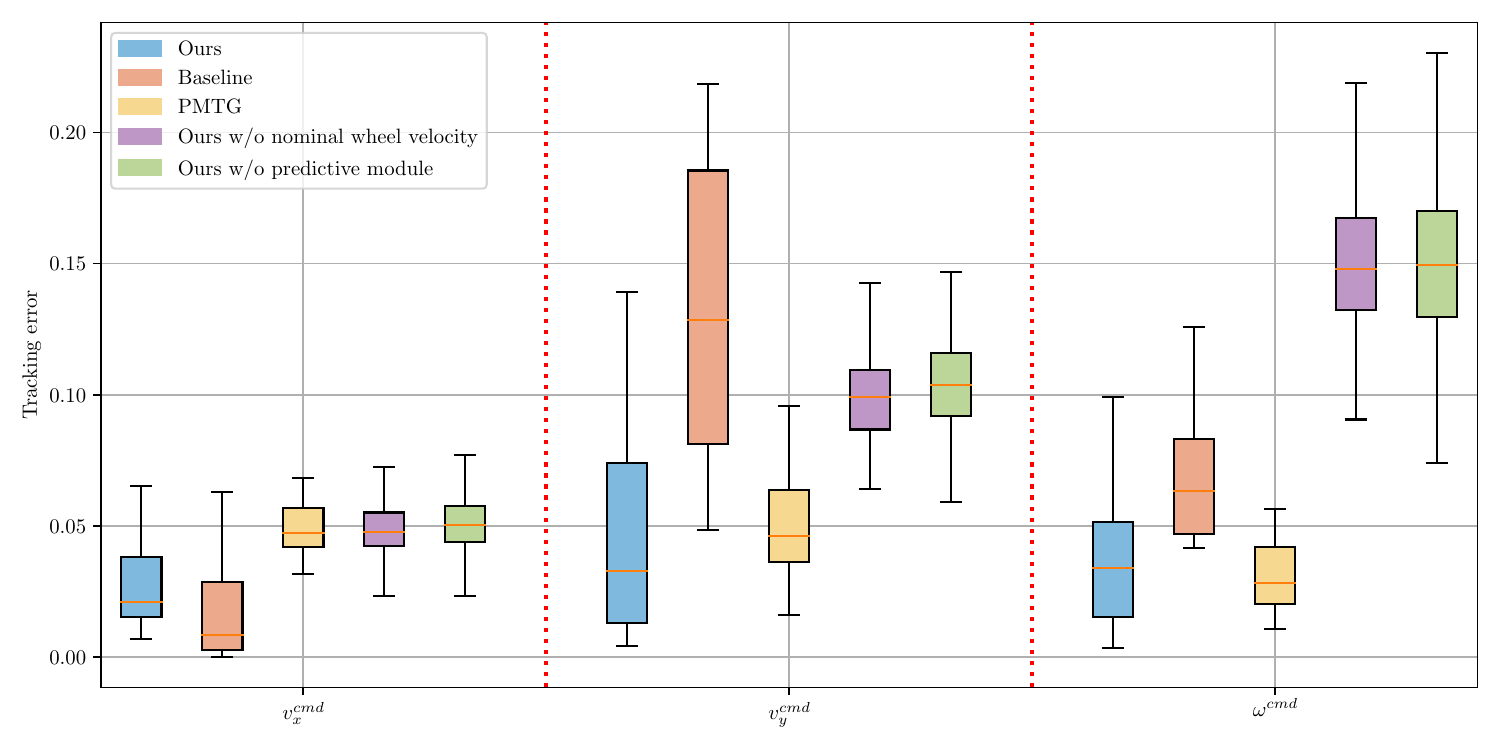}
        \includegraphics[width=0.45\textwidth]{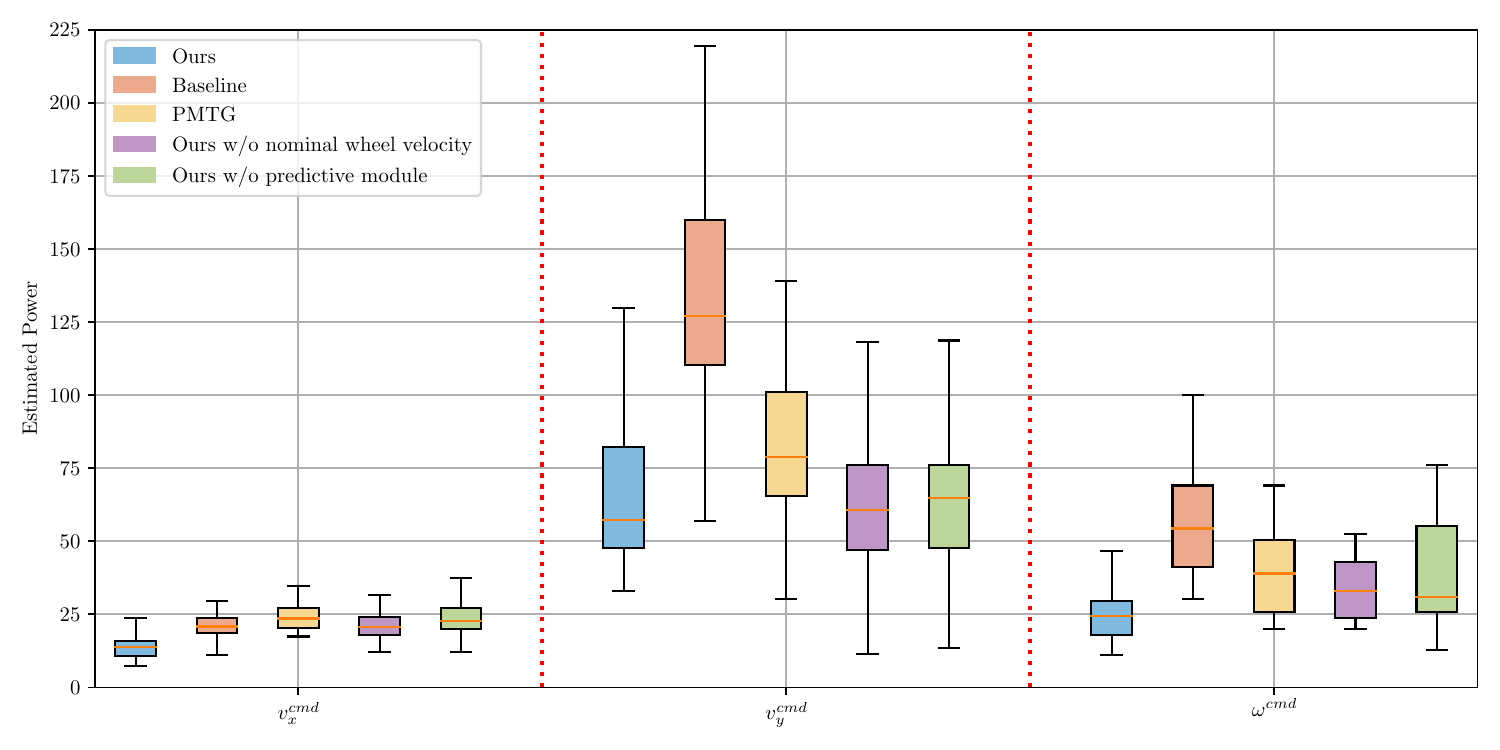}
        \captionsetup{font=footnotesize}
        \caption{Commanded velocity tracking error and estimated power for three tasks. We respectively sample $v_x^{\text{cmd}}\in[-1,1]$m/s, $v_y^{\text{cmd}}\in[-0.7, 0.7]$m/s and $\omega^{\text{cmd}}\in[-0.7, 0.7]$rad/s for each task.}
        \label{fig:velocity_tracking}
    \end{figure}

    \subsection{Simulation Experiment}
    \begin{table*}[!t]
        \caption{Ablation study on robustness to disturbances. The percentage of successful recovery from the disturbance is reported.}
        \label{tab:my-table}
        \centering
        \begin{tabular}{cclcccc}
            \hline
            \hline
        \multicolumn{1}{l}{Max. push} & Parameter Setting & \multicolumn{1}{c}{Ours}   & Baseline & PMTG & Ours w/o nominal wheel velocity & Ours w/o switching \\
        \hline
        \multirow{4}{*}{$\Delta v_{\max}^{\text{base}}$=0.5m/s}         & $v_x^{\text{cmd}}=v_y^{\text{cmd}}=\omega^{\text{cmd}}=0$& \multicolumn{1}{c}{\bf{100\%}}  & 95.6\%   & 68.5\% & 96.5\%   & 84.3\%             \\
        & $v_x^{\text{cmd}}\in[-1, 1]$   & \multicolumn{1}{c}{\bf{100\%}}  & 84.8\%   & 67.8\% & 92.7\%   & 82.8\%             \\
        & $v_y^{\text{cmd}}\in[-1, 1]$   & \multicolumn{1}{c}{\bf{99.8\%}} &   45.3\%  & 61.2\% & 87.1\%   & 76.0\%             \\
        & $\omega^{\text{cmd}}\in[-1, 1]$ & \multicolumn{1}{c}{\bf{99.7\%}} & 47\%     & 60.4\% & 91.2\%   & 78.9\%             \\
        \hline
        \multirow{4}{*}{$\Delta v^{\text{base}}_{\max}$=0.7m/s}         & $v_x^{\text{cmd}}=v_y^{\text{cmd}}=\omega^{\text{cmd}}=0$& \bf{98.7\%}& 89.8\%   & 58.4\% & 86.7\%   & 63.4\%             \\
        & $v_x^{\text{cmd}}\in[-1, 1]$   & \bf{97.6\%} & 72.5\%   & 57.3\% & 84.2\%   & 61.8\%             \\
        & $v_y^{\text{cmd}}\in[-1, 1]$   & \bf{91.7\%}&  26.5\%  & 56.8\% & 79.7\%   & 57.6\%             \\
        & $\omega^{\text{cmd}}\in[-1, 1]$ & \bf{93.8\%}& 39.1\%   & 54.3\% & 84.9\%   & 58.9\%             \\
        \hline
        \end{tabular}
        \end{table*}
    \begin{figure*}[!t]
        \centering
        \includegraphics[width=0.9\textwidth]{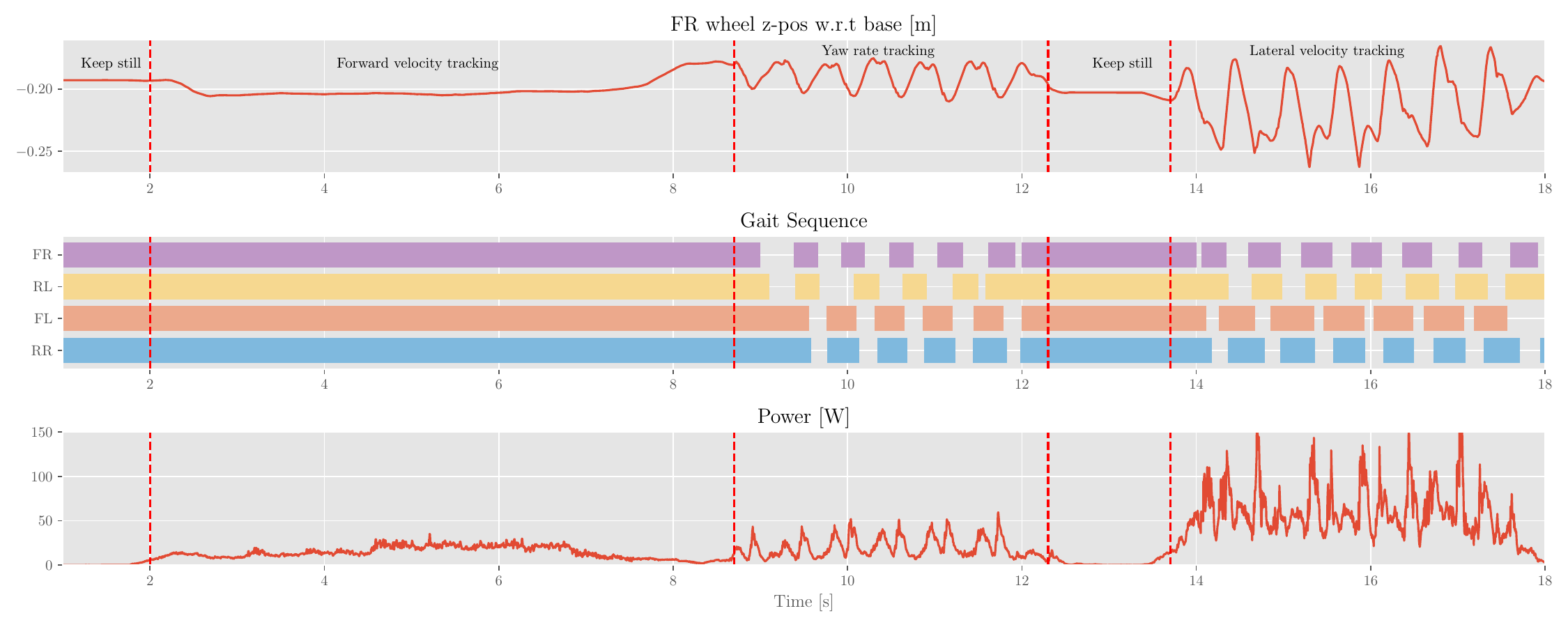}
        \caption{Commanded velocity tracking experiment. The vertical movement of the front right wheel, the gait sequence and the measured power are reported.}
        \label{fig:experiment}
    \end{figure*}

    {\bf{Commanded Velocity Tracking:}}
    We perform a sim-to-sim transfer evaluation to assess the performance of the proposed learning framework in the MuJoCo simulator~\cite{todorov2012mujoco}. The sim-to-sim code  along with the learned policy is available on the project website.

    To evaluate velocity tracking performance, we compute the mean squared tracking error (MSE) across three distinct tasks. In each task, one command from $v_x^{\text{cmd}}$, $v_y^{\text{cmd}}$, or $\omega^{\text{cmd}}$ is isolated, with the other two set to zero. Fig.~\ref{fig:velocity_tracking} presents the results as a bar plot, demonstrating that our framework consistently outperforms other methods in all tasks. The baseline policy, which lacks nominal gait guidance, fails to establish a stable policy for lateral velocity tracking, causing the robot to rely heavily on wheel movements for locomotion. Additionally, incorporating nominal wheel velocities significantly reduces tracking errors for forward velocity and yaw rate tasks. The learned switching between driving and trotting gaits further improves accuracy in lateral velocity and yaw rate tracking.

    {\bf{Energy Efficiency:}}
    Energy consumption is quantified by the robot's total power usage $p_{\text{cost}} = \sum_{i=1}^{16} |\tau_{i}|\cdot |q_{i}|$, where $\tau_i$ and $q_i$ represents the torque and the velocity of the $i$-th joint, respectively.

    In a setup similar to velocity tracking evaluation, each command is tested in isolation. We assess energy consumption over a 1-minute duration using 1000 robots. As shown in Fig.~\ref{fig:velocity_tracking}, energy consumption is minimized when only the forward velocity command is activated, owing to the reduced need for leg movement. Among the methods, PMTG exhibits the highest energy demand due to its reliance on a fixed nominal trotting gait. In contrast, our proposed framework demonstrates superior energy efficiency by optimizing the coordination of both leg and wheel motors. The selection of an appropriate nominal gait is key to minimizing energy, further highlighting the importance of nominal gait selection for optimal robot performance.

    {\bf{Robustness Analysis:}}
    To assess the robustness of the learned policy, we expose 1000 robots for a duration of 1 minute to simulated random pushes. These pushes are applied as random velocity perturbations $\delta v_{\text{max}}^{\text{base}}$ along the forward and lateral directions of the robot's trunk, with disturbances updated every 15 seconds. Maximum perturbation velocities of 0.5 m/s and 0.7 m/s are tested. The robot's robustness is evaluated by measuring the percentage of successful recoveries from the disturbances. As shown in Table~\ref{tab:my-table}, our framework significantly outperforms other methods across all scenarios, with a particularly notable advantage at the maximum push velocity of 0.9 m/s. The baseline policy, which fails to track lateral velocity, is not evaluated for robustness in this context. The introduction of the augmented nominal gait and automatic gait switching enhances the robot's resilience to external disturbances.

    \subsection{Real-World Experiment}
    The energy predictive module along with the learned policy are deployed on our wheeled quadrupedal robots, operating on the onboard Jetson Xavier NX at a frequency of 50 Hz. The desired joint angles of legs are tracked with PD controllers and the wheel motors are governed by PID controllers ensuring the tracking of desired wheel velocities. The instantaneous power consumption is estimated by the product of joint torques (proportional to motor currents for Unitree motors) and velocities.

    As illustrated in Fig.~\ref{fig:experiment}, the robot is commanded to sequentially perform tasks, including remaining stationary, moving forward, rotating and executing lateral movements. During stationary and forward movement phases, the wheel positions relative to the robot's base remain nearly constant. When adjusting for yaw rate and lateral velocity, the robot automatically switches to a trotting gait, relying primarily on leg movements. Interestingly, an increase in the legs' stepping height is observed during lateral movements compared with yaw rate tracking, highlighting the RL policy's adaptability.

    Energy consumption is monitored throughout the experiment and closely matched simulation results. The highest power occurs during lateral movements due to intensive leg motor usage, while the lowest energy consumption is recorded during stationary tasks, approaching zero due to minimal joint activity. The prediction module significantly reduces leg activity for forward movement and enhances the efficiency of wheeled-legged robots. For full experiment videos, please visit the project website.

    \section{Conclusion}\label{sec:conclusion}
    In this work, we present a hierarchical control framework that integrates a switching augmented nominal gait that is determined by an energy consumption prediction module with a learned residual RL policy, specifically designed to enhance the performance of wheeled quadrupedal robots. This framework enables the robot to automatically select the most efficient gait primitive and seamlessly transition between various gaits, dynamically adjusting its movement based on the task and environment. By coordinating both wheeled and legged locomotion, the system achieves adaptability, maintaining stability and maneuverability across a wide range of operating conditions, particularly when encountering external disturbances such as uneven surfaces or sudden pushes.

    \bibliographystyle{IEEEtran}
    \bibliography{./main,./IEEEabrv}

\end{document}